\documentclass[conference]{IEEEtran}
\IEEEoverridecommandlockouts
\usepackage{cite}
\usepackage{amsmath,amssymb,amsfonts}
\usepackage{algorithmic}
\usepackage{graphicx}
\usepackage{subfig}
\graphicspath{{./Images/}} 
\usepackage{textcomp}
\usepackage{xcolor}
\usepackage{multirow}
\usepackage{float}
\def\BibTeX{{\rm B\kern-.05em{\sc i\kern-.025em b}\kern-.08em
    T\kern-.1667em\lower.7ex\hbox{E}\kern-.125emX}}
\begin{document}

\title{On the Impact of Data Heterogeneity in Federated Learning Environments with Application to Healthcare Networks\\
}
\author{\IEEEauthorblockN{U. Milasheuski\textit{$^{1,2}$}, L. Barbieri\textit{$^{1}$}, B. Camajori Tedeschini\textit{$^{1}$}, M. Nicoli\textit{$^{1}$}, S. Savazzi\textit{$^{2}$}} \IEEEauthorblockA{\textit{$^{1}$Politecnico di Milano, Milan, Italy}, \textit{$^{2}$Consiglio Nazionale delle Ricerche, Milan, Italy}}}

\maketitle

\begin{abstract}
Federated Learning (FL) allows multiple privacy-sensitive applications to leverage their dataset for a global model construction without any disclosure of the information. One of those domains is healthcare, where groups of silos collaborate in order to generate a global predictor with improved accuracy and generalization. However, the inherent challenge lies in the high heterogeneity of medical data, necessitating sophisticated techniques for assessment and compensation. This paper presents a comprehensive exploration of the mathematical formalization and taxonomy of heterogeneity within FL environments, focusing on the intricacies of medical data. 
In particular, we address the evaluation and comparison of the most popular FL algorithms with respect to their ability to cope with quantity-based, feature and label distribution-based heterogeneity. 
The goal is to provide a quantitative evaluation of the impact of data heterogeneity in FL systems for healthcare networks as well as a guideline on FL algorithm selection. Our research extends beyond existing studies by benchmarking seven of the most common FL algorithms against the unique challenges posed by medical data use-cases. The paper targets the prediction of the risk of stroke recurrence through a set of tabular clinical reports collected by different federated hospital silos: data heterogeneity frequently encountered in this scenario and its impact on FL performance are discussed. 
\end{abstract}

\begin{IEEEkeywords}
Federated learning, healthcare networks, stroke prediction, distributed machine learning, heterogeneity
\end{IEEEkeywords}


\section{Introduction}
Machine Learning (ML) research has found its niche in various domains, including industry and healthcare systems \cite{ml4hlth}. Stroke, ranking second globally as a cause of death and third as a cause of death and disability combined, has drawn attention to ML applications, particularly in stroke recurrence prediction \cite{WSO}, \cite{stroke2}. Under this domain, ML models are typically trained on the data collected from multiple hospitals. However, sharing data across multiple hospitals, though effective, faces challenges due to strict privacy regulations. 

To overcome data-exchange issues, Federated Learning (FL) \cite{FEDAVG} allows sharing model weights while keeping data locally \cite{horFL}. Despite FL's privacy preservation, challenges arise from heterogeneity and non-identically Distributed (non-IID) data, impacting model convergence and performance across participants.

In healthcare-focused FL \cite{bernardo}, addressing data heterogeneity is crucial, as datasets collected from various hospitals may exhibit differences in patient demographics, medical equipment, and clinical practices. These disparities result in non-IID data distributions, where the data from different sources do not follow the same statistical patterns. 

Considering tabular datasets, i.e., to represent patient outcomes, the main components that introduce heterogeneity are the number of samples, the distribution of the labels and the distribution of the features, that is, the datasets may significantly vary in size, distribution of the targets and/or features. Based on this, the most recent state-of-the-art FL algorithms make their assumptions to try to compensate for the non-IID data on devices to optimize the process of federation.

State-of-the-art algorithms can be partitioned into algorithm-driven methods and data-driven methods \cite{fedtda}. While algorithm-driven methods mainly focus on designing loss functions, parameter aggregation strategies and other techniques to correct client drift in its local updates, data-driven methods aim to convert the data to IID data using augmentation techniques. Although many papers show that data-driven methods can improve the final performance of the model, like in \cite{yoshida2020hybridfl}, they pose privacy leakage risks in practical FL applications. Therefore, this study emphasizes algorithm-driven methods. Algorithms like FedAvg \cite{FEDAVG}, FedAdp \cite{FEDADP} and FedDkw\cite{FEDDKW}  introduce client coefficients to weigh contributions to the global model. FedProx \cite{FEDPROX} and FedDyn \cite{FEDDYN} penalize local losses to ensure consistent updates across devices. In contrast to the previous algorithms, FedNova \cite{FEDNOVA} performs gradient normalization for fairness during aggregation, while SCAFFOLD \cite{SCAFFOLD} introduces control variates for updates correction.

First investigations on a subset of algorithms in works like \cite{NOISE} and \cite{rev} focused on computer vision and time series tasks. These studies presented a lack of extensive experimental analysis in the healthcare domain for a proper understanding of their pros and cons. It emerges thereby a necessity to evaluate those algorithms under different non-IID scenarios specific to healthcare applications.

\begin{figure*}
\centerline{\includegraphics[width=0.85\textwidth]{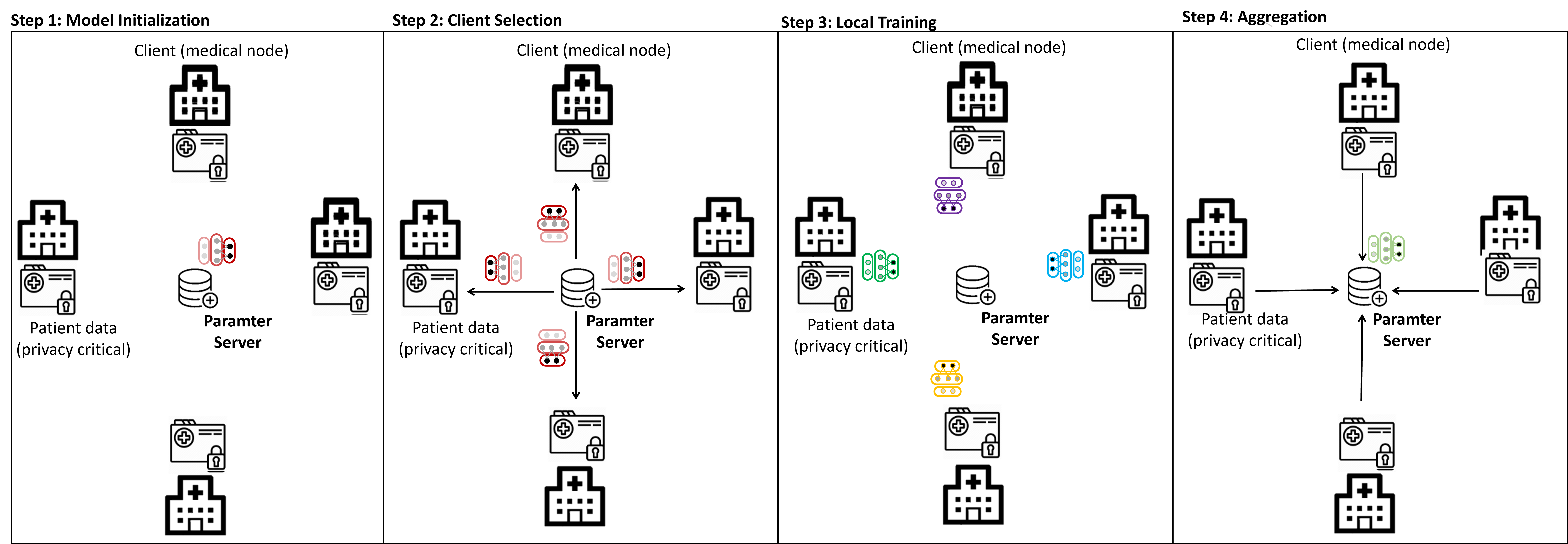}}
    \caption{General steps of a Federated Learning process.}
\label{fig:fl_process}
\end{figure*}

\textbf{Contributions.} 
In this paper, we first discuss a taxonomy of the main heterogeneity types for medical tabular data. The discussion includes methodologies used to simulate data heterogeneity in FL environments and, finally, a model to address data heterogeneity in real-world FL application settings. The paper also describes the design of a real-time FL networking system based on MQTT protocol and integrates seven state-of-the-art FL algorithms. Benchmarking of the algorithms was carried out for varying heterogeneity setups that are proposed to verify the sensitivity against quantity, label and feature skew. The study focuses on stroke prediction using a public dataset, aiming to offer insights for algorithm selection in various heterogeneity scenarios for tabular medical datasets.

The paper is organized as follows. Sect. II discusses the FL optimization. Sect. III describes the data heterogeneity in medical applications. Sect. IV discusses the adopted FL approach and algorithms. Sect. V analyzes the considered case study and performance assessment in various heterogeneous setups. Conclusions and final remarks are in Sect. VI.
\section{Distributed Optimization Modelling}

Let us consider a dataset $\mathcal{D}$ of size $|\mathcal{D}|$ distributed disjointly over the $K$ clients such that $\bigcup_{i=1}^{K} \mathcal{D}_i = \mathcal{D}$. The local datasets, denoted as $\mathcal{D}_i$ with size $|\mathcal{D}_i|$, is comprised of samples $\mathcal{D}_i=\{ (\boldsymbol{x}_i, y_i) \}$, where $\boldsymbol{x}_i, y_i$ represent feature columns and labels of client $i$ respectively. While the goal of the classic centralized approach is the minimization of the loss function w.r.t. the model parameters $w$ over all the data samples, FL involves the minimization of an average of the loss functions, each computed over data locally collected by individual medical sites w.r.t. $\boldsymbol{w}$. The problem is defined as follows:
\begin{equation}
    \min_{\boldsymbol{w} \in \mathbb{R}^d} L(w)=\min_{w \in \mathbb{R}^d}\sum_{i=1}^{K}p_iL_i(\boldsymbol{w}),
    \label{fl_opt}
\end{equation}
where  $L_i(w)=\frac{1}{|\mathcal{D}_i|}\sum_{(\boldsymbol{x_i},y_i) \in \mathcal{D}_i}l(\boldsymbol{w}, (\boldsymbol{x}_i, y_i))$ and with the weights assigned according to the size of a local dataset as $p_i = \frac{|\mathcal{D}_i|}{\sum_{i=1}^{K}|\mathcal{D}_i|}$.
This problem can be solved using Gradient Descent (GD) \cite{FEDAVG}, which involves iteratively computing 
\begin{equation}
    \boldsymbol{w}^{(t+1)}_{i}=\boldsymbol{w}_{i}^{(t)}-\eta_t\nabla L_{i}(\boldsymbol{w}^{(t)})
    \label{loc}
\end{equation}
for $t=0,1,2,...,T$, where $\eta_t$ is the learning rate. Given the  Eq. \eqref{fl_opt} with number of participants $S^{(t)}\le K$ for round $t$, the update rule becomes $\boldsymbol{w}^{(t+1)}=\boldsymbol{w}^{(t)}-\eta_t \frac{1}{|S^(t)|}\sum_{i \in S^(t)}p_i\nabla L_i(\boldsymbol{w}^{(t)})$. Considering large datasets, Stochastic Gradient Decent (SGD) is applied in practice. 
Besides initialization and client selection, as described in Fig. \ref{fig:fl_process}, the FL process consists of a local training stage, performed by clients according to Eq. \ref{loc}, followed by an aggregation stage, implemented by a Parameter Server (PS). During the aggregation, the PS produces a global model $w$ by solving the problem (Eq. \ref{fl_opt}) and using the model parameters $\boldsymbol{w}_{i}^{t}$ obtained from the clients. Aggregation and local training form a FL round and are repeated until convergence or a termination criterion is met.

\section{Heterogeneities in Medical Applications}

\begin{figure}
\centerline{\includegraphics[width=0.26\textwidth]{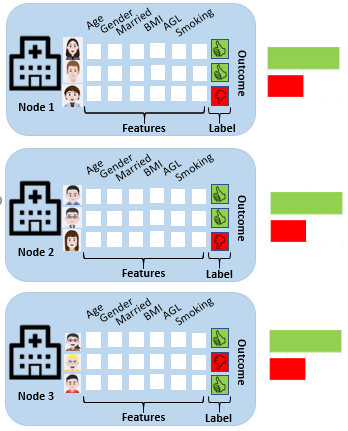}}
    \caption{An example of IID medical tabular datasets.}
\label{IID data}
\end{figure}

A common FL architecture in healthcare is horizontal federation \cite{horFL}. Each FL node collects data, namely clinical outcomes or reports from a subset of patients, while the reports consist of a common set of features or clinical variables. 


Numerous works, including \cite{10001718}, explore challenges arising from the unrealistic Independent and Identically Distributed (IID) assumption in FL. For a tabular medical dataset, illustrated in Fig. \ref{IID data}, green labels represent stroke patients (positive class), and red labels denote healthy patients (negative class). The histograms assume roughly equal sample numbers and consistent label distribution for IID data, but this assumption is often impractical. Despite efforts in works like \cite{FEDPROX} and \cite{SCAFFOLD} to address heterogeneity, this remains a critical and open problem, particularly in healthcare data scenarios.


Heterogeneity in medical data can arise for various reasons, including variations in data collection methods across institutions, missing data, and a significant source—patient diversity, encompassing demographic factors and clinical characteristics variability. Electronic health records from different institutions exhibit disparities in data formats and coding standards, complicating diagnosis comparison. Medical imaging data, including MRI scans and X-rays, vary in resolution and quality due to differences in equipment and protocols. Patient-reported outcomes introduce subjectivity, as individual interpretations and survey instruments vary. Addressing these heterogeneous data sources is vital for accurate analysis and informed decision-making in healthcare research and practice. For non-IID data we have that for client $i$ and $j$  the probabilities of drawing a data sample are $P(\boldsymbol{x}_i, y_i) \neq P(\boldsymbol{x}_j, y_j)$. Let the joint probability be rewritten as 

\begin{equation}
    P(\boldsymbol{x}_i, y_i)=P(y_i|\boldsymbol{x}_i)P(\boldsymbol{x}_i)=P(\boldsymbol{x}_i|y_i)P(y_i),
\end{equation}
heterogeneity might impact either the conditional distributions $P(\boldsymbol{x}_i|y_i)$ $P(y_i|\boldsymbol{x}_i)$, or the marginal ones $P(\boldsymbol{x}_i)$ and $P(y_i)$, giving rise to different data imbalance situations.

In the following, we consider tabular datasets representing typical clinical reports in medical applications. We classify the primary types of heterogeneity into three groups: \textit{label distribution skew}, \textit{quantity/sample distribution skew}, and \textit{feature distribution skew}.

\subsection{Label distribution skew}

\begin{figure*}
    \centering
    \subfloat[Label distribution skew. \label{Label distribution skew}]{
        \includegraphics[width=0.22\textwidth, height=4cm]{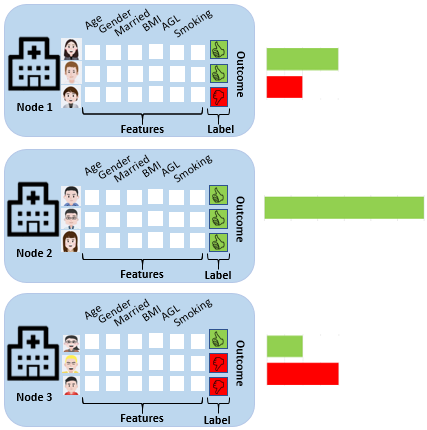}
    }
    \subfloat[Quantity skew. \label{Quantity skew}]{
        \includegraphics[width=0.22\textwidth, height=4cm]{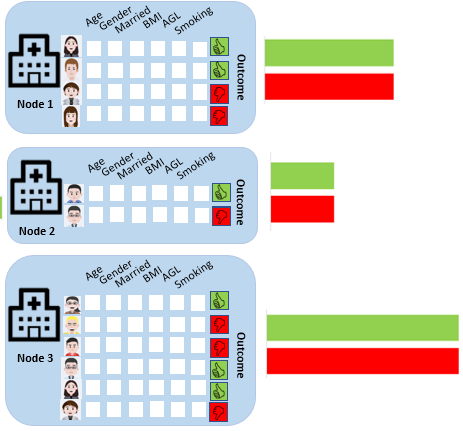}
    }
     \subfloat[Feature distribution skew \\ for a correlated feature. \label{Feature distribution skew 1}]{
        \includegraphics[width=0.22\textwidth, height=4cm]{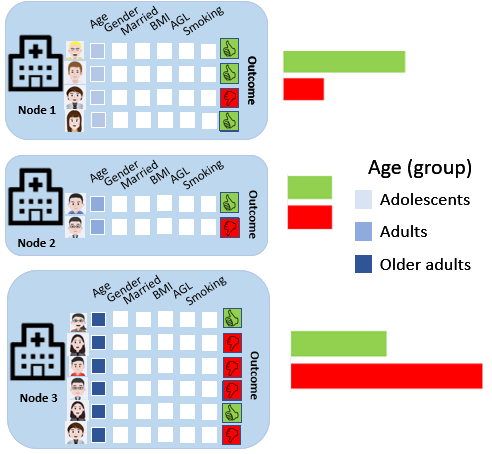}
    }
     \subfloat[Feature distribution skew \\ for an uncorrelated feature. \label{Feature distribution skew 2}]{
        \includegraphics[width=0.22\textwidth, height=4cm]{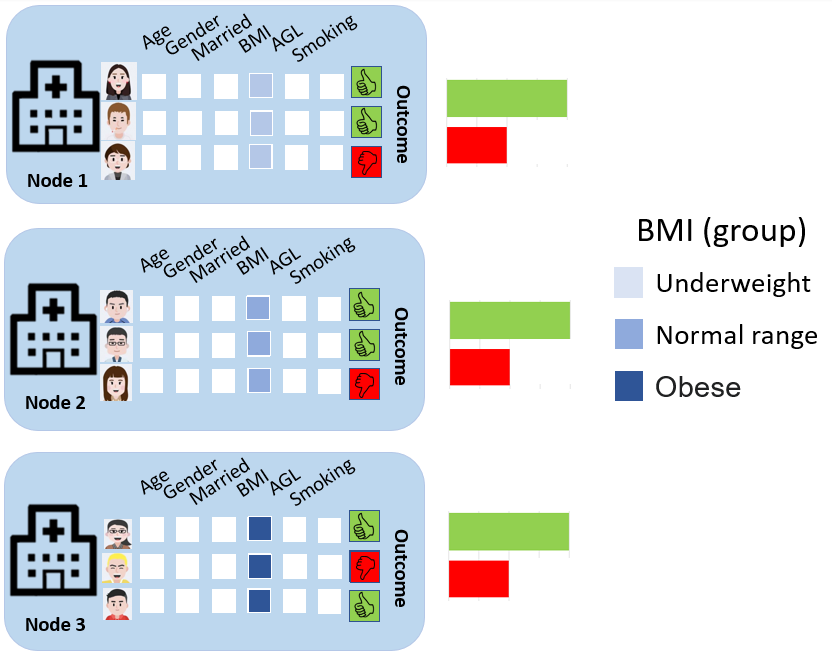}
    }
    \caption{Heterogeneity in medical tabular datasets data: label (a), quantity (b) and feature (c,d) distribution skew examples}
    \label{fig:squadre_genova}
\end{figure*}

For label distribution skew, patient numbers vary consistently across parties, while the class ratio in local datasets may differ significantly. This affects the marginals $P(y_i)$, so that $P(y_i) \neq P(y_j)$, while $P(\boldsymbol{x}_i | y_i)=P(\boldsymbol{x}_j | y_j) \ \forall i, j$. In Fig.~\ref{Label distribution skew}, the right histogram shows nodes 1 and 2 dominated by the positive class, and node 3 primarily composed of the negative class. This scenario might occur if a specific hospital provides superior treatment, resulting in better patient outcomes.

\subsection{Quantity skew}
 Regarding quantity skew, the setup involves nodes with the same class proportions per node, but the number of samples per node differs arbitrarily, i.e., $P(\boldsymbol{x}_i, y_i)=P(\boldsymbol{x}_j, y_j), \ |\mathcal{D}_i| \neq |\mathcal{D}_j| \ \forall i, j$. Figure~\ref{Quantity skew} illustrates an example where node 3 has significantly more patients than nodes 1 and 2, allowing it to perform more update steps during local training. For example, population density variations across locations might result in larger sample sizes for hospitals in denser areas compared to others.

 \subsection{Feature distribution skew}
The last heterogeneity type is feature distribution skew, where marginal distributions may vary across clients, i.e., $P(\boldsymbol{x_i}) \neq P(\boldsymbol{x_j}), \ P(y_i | \boldsymbol{x_i})=P(y_j | \boldsymbol{x_j}) \ \forall i, j$ (Fig.~\ref{Feature distribution skew 1} and \ref{Feature distribution skew 2}). In contrast to \cite{Advanced} and \cite{NOISE} which consider image datasets, preserving $P(y_i | \boldsymbol{x}_i)$ in targeted tabular data is not guaranteed and depends on feature-label correlation. Specific features, whether numerical or categorical, have non-uniform distributions, leading to sample imbalances on each node, as highlighted in Fig.~\ref{Feature distribution skew 2}, along with non-identical feature distributions within nodes. The blue color in the figure indicates a specific feature group, which results in limited variation within a federated node. 
 For instance, \textit{Age} feature skew (Fig. \ref{Feature distribution skew 1}) corresponds to cases where hospital sites observe medical records from a specific age group (young, middle-aged, or elderly). Assuming age strongly correlates with the probability of stroke, a hospital with more elderly patient reports likely has more stroke cases than other federated nodes. In other words, since elderly individuals are more likely to undergo medical treatment, their proportion would be larger than other age groups.
However, features like \textit{Body Mass Index} (\textit{BMI}), as depicted in Fig. \ref{Feature distribution skew 2}, may have low correlation with the overall label distribution (considering all reports). While this may be true, there could be situations where the feature within its range/categories on a node is highly correlated with the label distribution observed in local data. This implies that the client will rely on the feature differently. Similar reasoning can potentially apply to other numerical and categorical clinical features.
\section{MQTT-based FL Design}
This section outlines a FL system utilizing the MQTT protocol for model federation as proposed in \cite{bernardo}, and selected FL algorithms: on-device, on-PS, and hybrid algorithms. Additionally, it discusses methodologies for simulating heterogeneous environments.

\subsection{Communication Protocol}
The MQTT protocol facilitates message exchange between clients and the PS via a broker \cite{bernardo}. Utilizing the publish-subscribe pattern, devices publish trained models on the PS's topic for aggregation, while the PS sends the global model on $C$ topics representing each client. Additionally, a topic is reserved for individual client communication with the PS. Once the PS selects FL round participants, it publishes messages on each client's topic. After completing local updates, devices publish their models on the PS's topic. Quality of Service (QoS) is set to 2 for message receipt without repetitions, and the retain flag is set to \textit{True} for reliable message delivery in FL operations.

Excluding the unmodified MQTT packet header, the client's message payload includes: 1) client identifier, 2) federated round, 3) trainable parameters of the model (for each neural network layer), 4) metrics on the validation dataset (accuracy and loss), and 5) a boolean variable indicating local training completion. The PS payload shares a similar structure and includes additional algorithm-related parameters (see Sec. \ref{algs}), such as control variates for SCAFFOLD, training data distribution for FedDkw, etc.

\subsection{Selected Algorithms} \label{algs}
In a PS setup, aggregating algorithm-based methods can be categorized by their primary location of intelligence: on-PS for main steps during aggregation, on-client for local optimization, and hybrid for both local training and aggregation.

Starting with \textit{on-PS} algorithms, FedAvg \cite{FEDAVG} is a fundamental FL algorithm that aggregates parameters in proportion to the number of samples per client after optimization. FedAdp \cite{FEDADP} adjusts model weighting by estimating differences between local and global model gradients, measured through the angle between them. Another benchmarked algorithm, FedDkw \cite{FEDDKW}, employs Kullback–Leibler (KL) divergence \cite{KL} to compare the distribution of each device's training data to the global data distribution. However, a drawback is that, even though client data remains on clients, the data distribution is disclosed during the federation process, potentially revealing patterns or trends that could lead to the identification of individuals or specific health conditions.

In the \textit{on-device} category, FedProx \cite{FEDPROX} is chosen. This algorithm employs Euclidean distance between local and global models for regularization, penalizing divergence during local training. This straightforward regularization is particularly relevant for cross-device FL, as it avoids extra parameter transmission and doesn't introduce additional computational or communication complexities compared to basic FedAvg.
For \textit{hybrid} algorithms, we have chosen SCAFFOLD \cite{SCAFFOLD}, FedDyn \cite{FEDDYN}, and FedNova \cite{FEDNOVA}. SCAFFOLD, applicable in the cross-silo setting, utilizes control variates stored with each client to estimate the gradient of the loss with respect to the client's local data. The server maintains the average of all client states as its control variate, shared with selected clients in each round. FedDyn, similarly to FedProx, adjusts the local objective function with a dynamically updated regularizer consisting of the same term as FedProx and the dot product between local gradients and the global model. During aggregation, the PS tracks parameters called server state, influencing the model to debias it. While this regularization doesn't introduce extra communication like SCAFFOLD, it requires clients to maintain states or memory across rounds. In contrast, FedNova aims to eliminate inconsistency between local updates by normalizing them with the number of local steps. This method acts as a new weighting scheme, assigning lower weights to clients with more local steps, potentially preventing them from pushing the aggregated model towards their local minima after the aggregation step.

\subsection{Simulation of Heterogeneous
Data}
To synthesize a data distribution in a heterogeneous way we used a Dirichlet distribution with Probability Density Function (PDF) defined as follows:

\begin{equation}
    \label{eq:dir}
   \text{Dir}(\boldsymbol{p}|\boldsymbol{\alpha}) = \frac{1}{B(\boldsymbol{\alpha})} \prod_{i=1}^{K} p_i^{\alpha_i - 1},
\end{equation}
where $\boldsymbol{p}$ is the $K \times 1$ vector of probabilities representing a point,  $K$ is the number of variables, $\boldsymbol{\alpha}$ is the $K \times 1$ vector of concentration parameters, while $B(\boldsymbol{\alpha})$ is the multivariate beta function defined as $B(\boldsymbol{\alpha}) = \frac{\prod_{i=1}^{K}\Gamma(\alpha_i)}{\Gamma(\sum_{i=1}^{K} \alpha_i)}$, with $\Gamma(x)$ denoting the gamma function, $p_i$ is the $i$-th element of $\boldsymbol{p}$.

In particular, we use a \textit{symmetrical Dirichlet distribution} to simulate non-IID data by distributing the samples in each class unequally between clients, in proportions sampled from a symmetrical K-dimensional Dirichlet distribution with $\boldsymbol{\alpha}=\alpha\cdot\mathbf{1}$, while $\boldsymbol{1}$ denotes the unitary vector $K \times 1$, as in  \cite{hsu2019measuring}. 

The $\alpha$ hyperparameter can be tuned to control how unequally the data classes are divided between clients. Large values of $\alpha$ lead to a low variance in proportion, resulting in more equal splits between clients (low heterogeneity), whereas low $\alpha$ increases the variance of the proportions, leading to a more non-IID data split (high heterogeneity). 

For data partitioning, we implement the following heterogeneity scenarios:
\begin{itemize}
    \item \textbf{Quantity Skew}:
    By applying the Dirichlet distribution for each client we can vary the assigned number of samples per device as $|D_i| = \lfloor p_i|D| \rfloor$, where $|D|$ is the total amount of data samples available, $p_i \in \mathbf{p}$ is the probability of device $i$ having a sample. The label distribution for each of the devices is identical.

\item \textbf{Label Distribution Skew}:
 For every label $j$, given $\alpha_i$ for each client $i$, we sample $\boldsymbol{q^j} \sim Dir(\boldsymbol{p}|\boldsymbol{\alpha})$ of size $K \times 1$ and assign to a client the corresponding number of samples of a specific class $|D_{i}^j| = \lfloor q_{i}^{j}|D^j| \rfloor$, where $|D_{i}^j|$ - number of samples of class $j$ is assigned to the client $i$. This procedure is repeated for every class to perform the partitioning amongst the clients. The number of samples per device is set to be the same.

\item \textbf{Feature Skew}:
In order to partition the dataset properly, it was decided to select the most and the least correlated features to the label. After that, we decided to discretize the range of the feature in two different ways, either by dividing it into $K$ equal ranges, where $K$ is the number of clients, or by partitioning it such that the number of samples per client is approximately the same.

\end{itemize}

\section{Impact of Heterogeneities on the Federation}
This section benchmarks the current state-of-the-art FL algorithms, evaluating their performance on real tabular data from publicly available health records to identify stroke risk. The objective is to gain insight into the algorithms' ability to handle heterogeneity and offer general guidelines for algorithm selection in the proposed case study.

\begin{figure*}
    \centering
    \subfloat[Label, $\alpha=0.1$ \label{fig:label_imbalance_alpha=0.1}]{
        \includegraphics[width=0.23\textwidth]{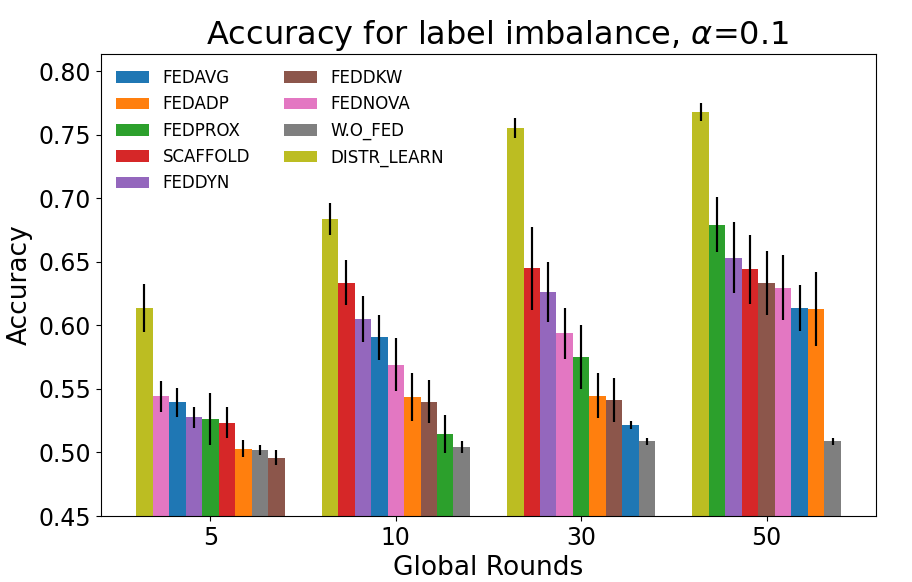}
    }
    \subfloat[Label, $\alpha=0.3$ \label{fig:label_imbalance_alpha=0.3}]{
        \includegraphics[width=0.23\textwidth]{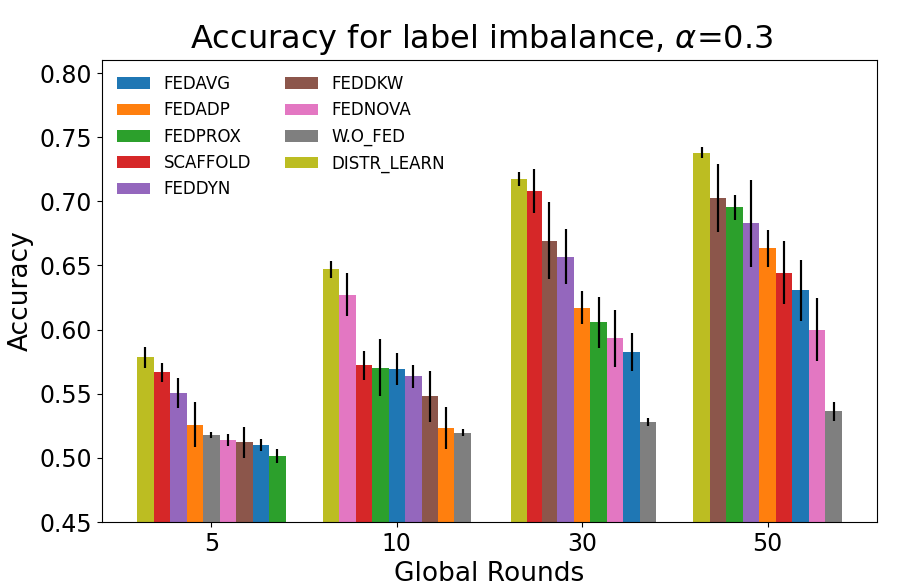}
    }
    \subfloat[Quantity, $\alpha=0.1$ \label{fig:samples_imbalance_alpha=0.1}]{
        \includegraphics[width=0.23\textwidth]{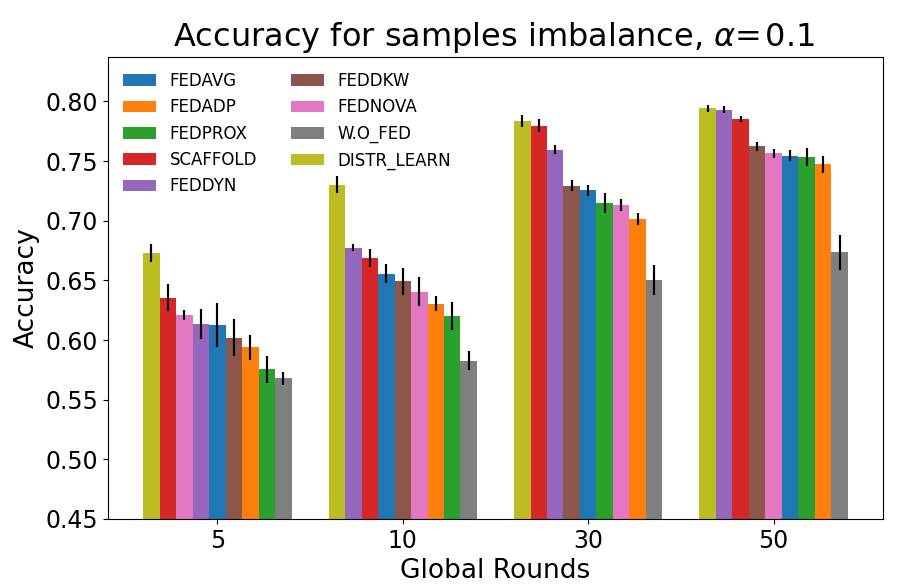}
    }
    \subfloat[Quantity, $\alpha=0.3$ \label{fig:samples_imbalance_alpha=0.3}]{
        \includegraphics[width=0.23\textwidth]{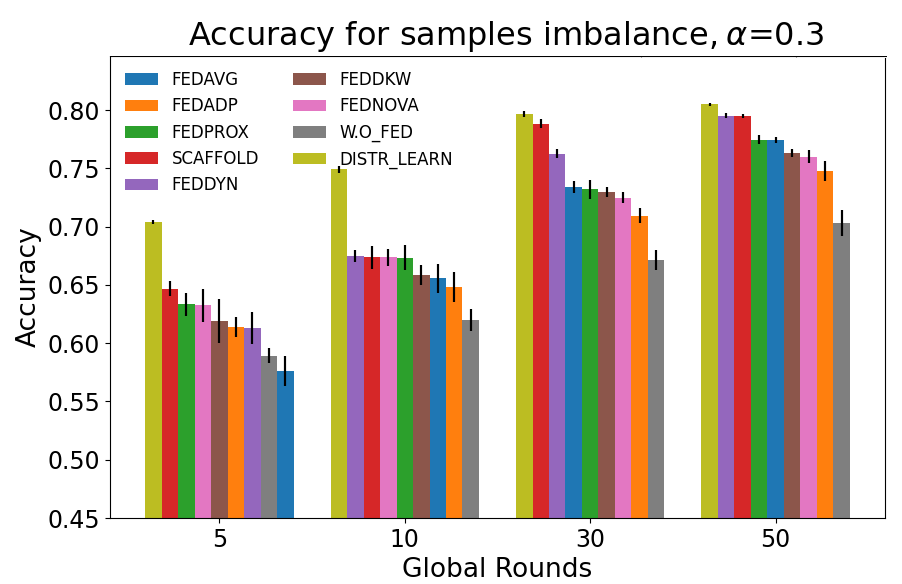}
    }

    \subfloat[Feat., ev. samp., Age
    \label{fig:feat_skew_age_samp}]{\includegraphics[width=0.23\textwidth]{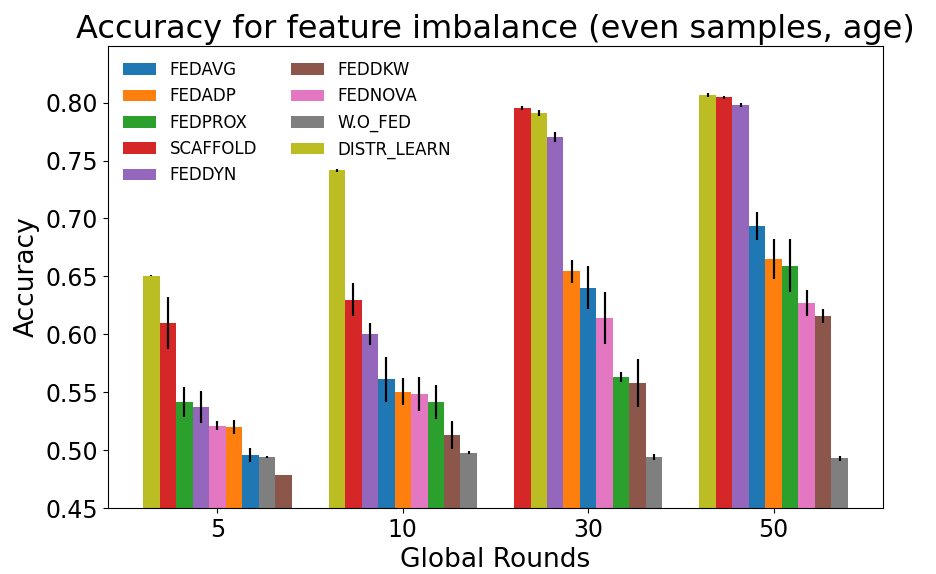}
    } 
    \subfloat[Feat., ev. int., Age
    \label{fig:feat_skew_age_int}]{\includegraphics[width=0.23\textwidth]{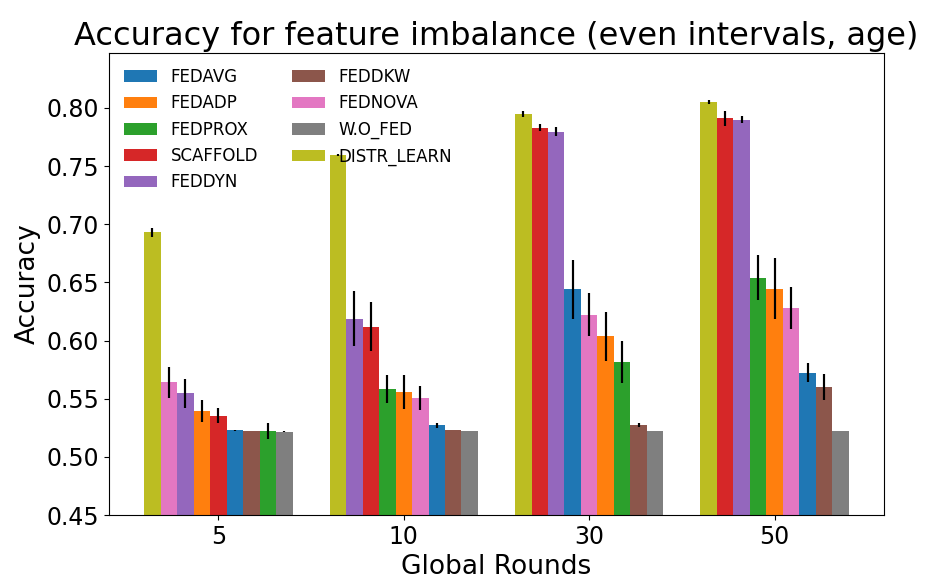}
    }
    \subfloat[Feat., ev. samp., BMI
    \label{fig:feat_skew_bmi_samp}]{\includegraphics[width=0.23\textwidth]{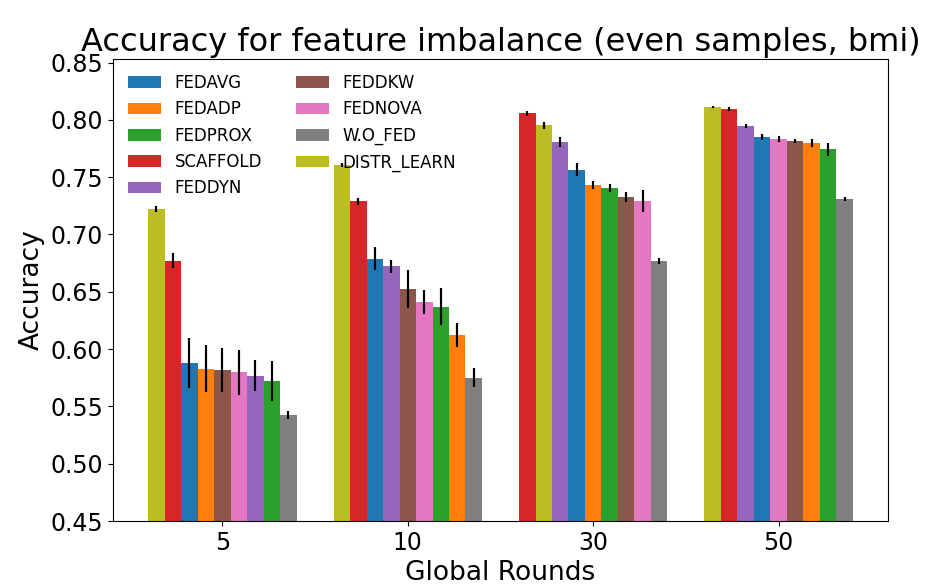}
    } 
    \subfloat[Feat., ev. int., BMI
    \label{fig:feat_skew_bmi_int}]{\includegraphics[width=0.23\textwidth]{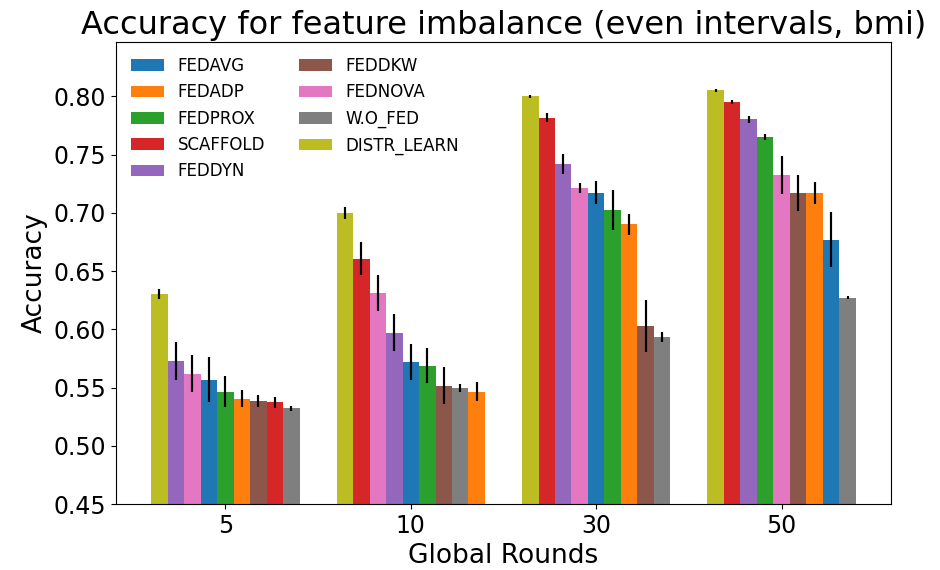}
    }
    
    \caption{FL algorithms evaluation on stroke dataset: a), b) - Label distribution skew; c), d) - Quantity skew, e), f) - Feature distribution skew for feature Age; g), h) - Feature distribution skew for feature BMI.}
\end{figure*}

    

\begin{table*}[]
\begin{tabular}{cc|cc|cc|cccc|}
\cline{3-10}
                                          &               & \multicolumn{2}{c|}{Label}                            & \multicolumn{2}{c|}{Quantity}                         & \multicolumn{4}{c|}{Feature}                                                                                                     \\ \hline
\multicolumn{1}{|c|}{Round}               & Alg           & \multicolumn{1}{c|}{$\alpha = 0.1$} & $\alpha = 0.3$  & \multicolumn{1}{c|}{$\alpha = 0.1$} & $\alpha = 0.3$  & \multicolumn{1}{c|}{BMI, ev, samp.} & \multicolumn{1}{c|}{BMI, ev. int.} & \multicolumn{1}{c|}{Age, ev. samp.} & Age, ev, int.   \\ \hline
\multicolumn{1}{|c|}{\multirow{9}{*}{30}} & FEDAVG        & 0.5216                              & 0.5827          & 0.7254                              & 0.7336          & 0.7566                              & 0.7174                             & 0.6403                              & 0.6439          \\ \cline{2-2}
\multicolumn{1}{|c|}{}                    & FEDADP        & 0.5447                              & 0.6171          & 0.7012                              & 0.7094          & 0.7431                              & 0.6901                             & 0.6544                              & 0.6036          \\ \cline{2-2}
\multicolumn{1}{|c|}{}                    & FEDPROX       & 0.5752                              & 0.6056          & 0.7147                              & 0.732           & 0.7406                              & 0.7026                             & 0.5633                              & 0.5812          \\ \cline{2-2}
\multicolumn{1}{|c|}{}                    & SCAFFOLD      & \textbf{0.6448}                     & \textbf{0.7126} & \textbf{0.7797}                     & \textbf{0.7881} & \textbf{0.8055}                     & \textbf{0.7816}                    & \textbf{0.7953}                     & \textbf{0.7828} \\ \cline{2-2}
\multicolumn{1}{|c|}{}                    & FEDDYN        & 0.6266                              & 0.657           & 0.7595                              & 0.7626          & 0.7806                              & 0.7418                             & 0.7704                              & 0.7793          \\ \cline{2-2}
\multicolumn{1}{|c|}{}                    & FEDDKW        & 0.5413                              & 0.6694          & 0.7295                              & 0.7299          & 0.7325                              & 0.6029                             & 0.5579                              & 0.5272          \\ \cline{2-2} 
\multicolumn{1}{|c|}{}                    & FEDNOVA       & 0.5938                              & 0.5931          & 0.7129                              & 0.7248          & 0.7294                            & 0.7215                              & 0.6144         & 0.6221          \\ \cline{2-2} 
\multicolumn{1}{|c|}{}                    & W.O. FED      & 0.5088                              & 0.528           & 0.6502                              & 0.6714          & 0.6767                              & 0.5937                             & 0.4941                              & 0.5219          \\ \cline{2-2}
\multicolumn{1}{|c|}{}                    & DISTR. LEARN. & 0.7554                              & 0.7175          & 0.7835                              & 0.7965          & 0.795                               & 0.8001                             & 0.7911                              & 0.7949          \\ \hline
\multicolumn{1}{|c|}{\multirow{9}{*}{50}} & FEDAVG        & 0.6138                              & 0.6306          & 0.7546                              & 0.7741          & 0.785                               & 0.6769                             & 0.6936                              & 0.5723          \\ \cline{2-2}
\multicolumn{1}{|c|}{}                    & FEDADP        & 0.6128                              & 0.6633          & 0.7472                              & 0.7479          & 0.7798                              & 0.7169                             & 0.6654                              & 0.6446          \\ \cline{2-2}
\multicolumn{1}{|c|}{}                    & FEDPROX       & \textbf{0.6792}                     & 0.6952          & 0.7533                              & 0.7745          & 0.7741                              & 0.7654                             & 0.6592                              & 0.6541          \\ \cline{2-2}
\multicolumn{1}{|c|}{}                    & SCAFFOLD      & 0.644                               & 0.6443          & 0.785                               & 0.7951          & \textbf{0.8094}                     & \textbf{0.795}                     & \textbf{0.8046}                     & \textbf{0.7908} \\ \cline{2-2}
\multicolumn{1}{|c|}{}                    & FEDDYN        & 0.6534                              & 0.6827          & \textbf{0.7931}                     & \textbf{0.7953} & 0.7946                              & 0.7805                             & 0.7977                              & 0.7899          \\ \cline{2-2}
\multicolumn{1}{|c|}{}                    & FEDDKW        & 0.6335                              & \textbf{0.7025} & 0.7624                              & 0.763           & 0.7816                              & 0.7172                             & 0.6158                              & 0.5602          \\ \cline{2-2}
\multicolumn{1}{|c|}{}                    & FEDNOVA       & 0.6298                              & 0.6             & 0.7565                              & 0.7598          & 0.7829                              & 0.7324                             & 0.6268                              & 0.6281          \\ \cline{2-2}
\multicolumn{1}{|c|}{}                    & W.O. FED      & 0.5088                              & 0.5364          & 0.6735                              & 0.7033          & 0.7312                              & 0.6273                             & 0.4928                              & 0.522           \\ \cline{2-2}
\multicolumn{1}{|c|}{}                    & DISTR. LEARN. & 0.7678                              & 0.7379          & 0.7943                              & 0.8048          & 0.8115                              & 0.8052                             & 0.8068                              & 0.8052          \\ \hline
\end{tabular}
\caption{Accuracy of the algorithms in heterogeneous setups for stroke prediction dataset}
\label{tab:STROKE_acc}
\end{table*}

\subsection{Dataset Description and Simulation Setup}


The proposed case study utilizes the publicly available Stroke Prediction dataset \cite{stroke-dataset}. This dataset shares common features with a real dataset and serves as a benchmark, for example in \cite{stroke_usage}. It consists of 5110 clinical reports, 12 clinical attributes and corresponding labels. Key attributes crucial for monitoring stroke recurrence risk include \textit{'Gender'}, \textit{'Age'}, \textit{'Hypertension'}, \textit{'Heart disease indicator'}, \textit{'Married status'}, \textit{'Work type'}, \textit{'Residence type'}, \textit{'Glucose level'}, \textit{'Body Mass Index (BMI)'}, and \textit{'Smoking status'}. The labels indicate the risk of stroke for the patient, currently taking binary values (future works will consider more complex cases).

Data preprocessing involved outlier removal using the Interquartile Range (IQR) and filling missing values with feature-specific mean values for each gender. Non-categorical clinical variables were normalized by subtracting the mean and dividing by the standard deviation, while categorical features underwent one-hot encoding. To address class imbalance, SMOTE \cite{SMOTE} was utilized for oversampling the minority class.
The predictive model in the FL setting is a deep neural network with three fully-connected layers, each followed by a dropout layer. Simulations assume $K=6$ clinical sites, with $S^{(t)}=2$ clients selected per round. Local training comprises $m=4$ epochs, with $b=3$ batches of size $B=50$ samples. The maximum samples per site are capped at $500$.

Two reference scenarios are considered for FL performance benchmarking:

\begin{enumerate}
    \item \textit{No federation}: clients independently use their datasets to train local models without federation. Performance is averaged over all clients;
  
\begin{table}[H]
\centering
\begin{tabular}{|c|c|}
\hline 
Level & Range of $\alpha$ \\
\hline
Extreme & $0 < \alpha < 0.1$ \\
\hline
High & $0.1 \leq \alpha < 0.3$ \\
\hline
\text{High/Medium} & $0.3 \leq \alpha < 0.5$ \\
\hline
Medium & $0.5 \leq \alpha < 0.7$ \\
\hline
Low & $0.7 \leq \alpha < 10$ \\
\hline
Homogeneous & $\alpha \geq 10$ \\
\hline
\end{tabular}
\caption{Heterogeneity levels for different concentration parameters.}
\label{tab:alpha_levels}
\end{table}

  \item  \textit{Distributed Learning}: data is transferred to a data center, which collects and manages the data for training. 
\end{enumerate} 
It is important to highlight, that these two benchmarks show the best and worst possible setups for comparison purposes only. Therefore, for instance, in Table \ref{tab:STROKE_acc}, they will not be considered during the discussion of the algorithms' performance.

In addition, we define $6$ different levels of heterogeneity by mapping the intervals of $\alpha$ shown in Table \ref{tab:alpha_levels}. 
These approximate ranges are based on the simulations performed on MNIST dataset, similar to simulations done in \cite{NOISE}, \cite{feddc} etc., and on the impact of the concentration parameter on the KL divergence of the Dirichlet and the uniform distributions.


\subsection{Performance Benchmarking: Label Distribution Skew}
The main results, depicted in Figs. \ref{fig:label_imbalance_alpha=0.1}-\ref{fig:label_imbalance_alpha=0.3}, illustrate accuracy (percentage of correct predictions) for various $\alpha$ values representing label imbalance. In scenarios with significant heterogeneity ($\alpha \leq 0.3$), all algorithms outperform the average performance of the non-federated scenario. FedProx excels in this context after a large number of rounds for medium/high and high heterogeneities.  SCAFFOLD initially outperforms the other algorithms for $\alpha=0.1$ but maintains the same accuracy over the rounds. For \textit{high/medium} heterogeneity, FedDkw shines due to the classification problem being binary and a smaller level of heterogeneity, with KL divergence built of only two bins. In \textit{high/medium} heterogeneity, SCAFFOLD, FedNova, and FedDyn enhance local gradients during aggregation, significantly improving performance. In cases of major dataset differences, simpler algorithms that penalize the local objective function provide the best results.

\subsection{Performance Benchmarking: Quantity skew}

    

Figures \ref{fig:samples_imbalance_alpha=0.1}-\ref{fig:samples_imbalance_alpha=0.3} display accuracy in the quantity skew setup for various heterogeneity levels. Quantity skew is less pronounced compared to label skew. For all heterogeneity levels, SCAFFOLD and FedDyn outperform, achieving over 77\% accuracy. As $\alpha$ increases, other algorithms tend to perform similarly, as the data maintains a fixed distribution on each client, and the average samples per client increase. The results highlight the ability of advanced hybrid algorithms to correct updates with persistent states, while algorithms relying on aggregation weights exhibit similar performance.

\subsection{Performance Benchmarking: Feature Distribution Skew} 
Feature skew arises from variations in patient demographics across hospitals. These variations impact model training. Based on an analysis of feature correlations with the target, we chose to partition data based on the feature corresponding to the minimum (\textit{BMI}) or maximum (\textit{Age}) Pearson correlation values. Two types of splits were performed: maintaining \textit{even intervals} within the feature range (split the feature range into equal non-overlapping windows) and \textit{even samples} (split the feature range into non-uniform non-overlapping windows with the same number of samples per window) to retain consistent sample counts per client. In the first approach, perspectives are limited to narrow feature subsets, potentially missing important patterns in other regions. This is crucial as nuances or trends may be concentrated in specific regions of the feature space. On the other hand, the second approach, maintaining even samples, is important to ensure that each client contributes an equal amount of information to the analysis. This prevents biases that may arise if certain clients are overrepresented or underrepresented in the dataset. 





Figures.~\ref{fig:feat_skew_age_samp} and \ref{fig:feat_skew_bmi_samp} show the accuracy for the partitioning into bins with \textit{even samples} of features \textit{Age} and \textit{BMI}respectively. For this case, except for FedDyn and SCAFFOLD, performances are similar for the less correlated feature. However, the \textit{Age} feature has a more pronounced impact on performance. This happens because when there is a skewed distribution of features, it indirectly leads to various other types of differences due to correlations. For instance, since stroke correlation with \textit{BMI} is low, \textit{even samples} partitioning equates to approximately IID data across clients, while for \textit{Age} \textit{even samples} partitioning leads to label distribution skew.

The results for partitioning features \textit{Age} and \textit{BMI} into \textit{even intervals} are highlighted in Figs. \ref{fig:feat_skew_age_int} and \ref{fig:feat_skew_bmi_int} respectively. Partitioning on \textit{Age} has a significantly noticeable impact on most of the algorithms. As before, SCAFFOLD and FedDyn consistently perform better than the remaining ones. \textit{Even intervals} partitioning for the least correlated feature is equivalent to sample imbalance with a small $\alpha$ value. In the case of partitioning the most correlated feature is equivalent to the combination of quantity and label skews. 

For both types of partitioning of the most correlated numerical feature, the most outperforming algorithms are the hybrid algorithms SCAFFOLD and FedDyn, outperforming others by over $12\%$. However, in the case of the least correlated feature, these algorithms provide a relatively small advantage compared to the other benchmarked algorithms.

\subsection{Guidelines for algorithm selection }
Table \ref{tab:STROKE_acc} provides the accuracy of the algorithms in diverse heterogeneous setups for a tabular stroke prediction dataset. Of the setups covered in the paper, the most impacting type of heterogeneity is label distribution skew. SCAFFOLD and FedDyn seemed promising, specifically for feature imbalance, with SCAFFOLD outperforming FedDyn, though demanding twice the parameters transmission, while FedDyn needed more computations. FedProx, FedNova, or FedAvg could be used when the devices have constraints on bandwidth or computational power. One of the main differences when comparing the algorithms with the computer vision tasks is the low dimensionality of the problem for the tabular data, which might be crucial for benchmarking. Overall, a-priori information on estimated heterogeneity levels is needed, prompting consideration of a common metric for its estimation. If no constraints exist, SCAFFOLD performs generally well for most of the cases, and if throughput is a concern, FedDyn is a promising alternative, unless there are limitations on keeping the state in memory or computations.
\section{Conclusions}
The paper explores the challenges of applying Federated Learning (FL) to heterogeneous medical tabular data, particularly in the context of stroke risk prediction. It identifies and categorizes various types of heterogeneity in healthcare data, such as variations in formats, tabular structures, and sources, as well as it provides a mathematical formalization for each of the proposed data imbalance scenario. The study then conducts a thorough benchmarking of seven state-of-the-art FL algorithms using publicly available stroke records.

Notably, the paper provides insights into the performance, communication resource usage, and learning round efficiency of the FL algorithms, offering a comprehensive evaluation tailored to the specific healthcare domain. The choice of a MQTT-based platform-as-a-service tool is also highlighted as a novel solution for managing communication and collaboration among distributed healthcare data sources.

Future work is outlined to compare neural networks with classification and regression tree models, such as random forests and XGBoost, in FL environments. Additionally, the exploration of the possibility of optimizing the ensemble of FL models will be done to enhance both performance and efficiency. Moreover, we acknowledge that the primary limitation lies in the dataset utilized. Our plan includes the acquisition of more realistic data to facilitate more practical simulations.

\bibliographystyle{IEEEtran}
\bibliography{biblio}










\end{document}